\documentclass[conference]{IEEEtran}
\IEEEoverridecommandlockouts

\usepackage{xcolor}
\usepackage{colortbl}
\definecolor{savegreen}{RGB}{34,139,34}

\usepackage{cite}
\usepackage{amsmath,amssymb,amsfonts}
\usepackage{graphicx}
\usepackage{algorithm2e}

\usepackage{textcomp}
\usepackage{xcolor}
\usepackage{graphicx}
\usepackage{comment}
\usepackage[bookmarks=false,hypertexnames=false]{hyperref}
\usepackage{amsmath}
\usepackage{xcolor}
\usepackage{color,soul}
\def\BibTeX{{\rm B\kern-.05em{\sc i\kern-.025em b}\kern-.08em
    T\kern-.1667em\lower.7ex\hbox{E}\kern-.125emX}}
\begin{document}

\title{FoveaDet: Software-Hardware Co-Design of Foveated CNN-Transformer for Object Detection\\}

                
        
\author{\IEEEauthorblockN{Deepak Mewada$^1$, Apporva Saraf$^2$, Ashwin Lele$^1$}
\IEEEauthorblockA{\textit{$^1$Centre for Machine Intelligence and Data Science (CMInDS), IIT Bombay, India}}
\IEEEauthorblockA{\textit{$^2$Department of Electrical Engineering, IIT Delhi, India}} 

{deepak.mewada@iitb.ac.in, ee1231164@iitd.ac.in, ashwin.lele@iitb.ac.in}}

\maketitle

\begin{abstract}
Deploying real-time object detection on power-constrained edge platforms is challenging as processing full-resolution frames is often computationally redundant. Inspired by biological foveation which assigns variable resolution based on visual saliency, we propose a novel software-hardware co-design framework called FoveaDet. On the software side, FoveaDet dynamically identifies salient regions to predict fixation points, followed by aggressive patch pruning in Vision Transformers (ViT) to discard irrelevant background data. To mitigate slowdowns from sequential patch generation, we implement optimized interpolation and patch selection algorithms. Evaluated on standard GPUs, this foveated approach reduces floating-point operations by 65.4\% to 92.2\% and accelerates inference by 4.76$\times$, while suffering a negligible mAP drop compared to dense baselines. To eliminate hardware bottlenecks, we complement the algorithm with a custom RISC-V core featuring specialized instructions to accelerate the foveated processing pipeline. FoveaDet provides a scalable, efficient architecture capable of unlocking advanced vision processing on severely constrained mobile platforms.

\end{abstract}

\begin{IEEEkeywords}
Vision Transformer, Foveation, Efficient Systems, Edge Computing, RISC-V Core
\footnote{Code : https://github.com/Deepakmewadaa/FoveaDet}
\end{IEEEkeywords}

\section{Introduction}

\begin{figure}[h!]
\centerline{\includegraphics[width=0.5\textwidth]{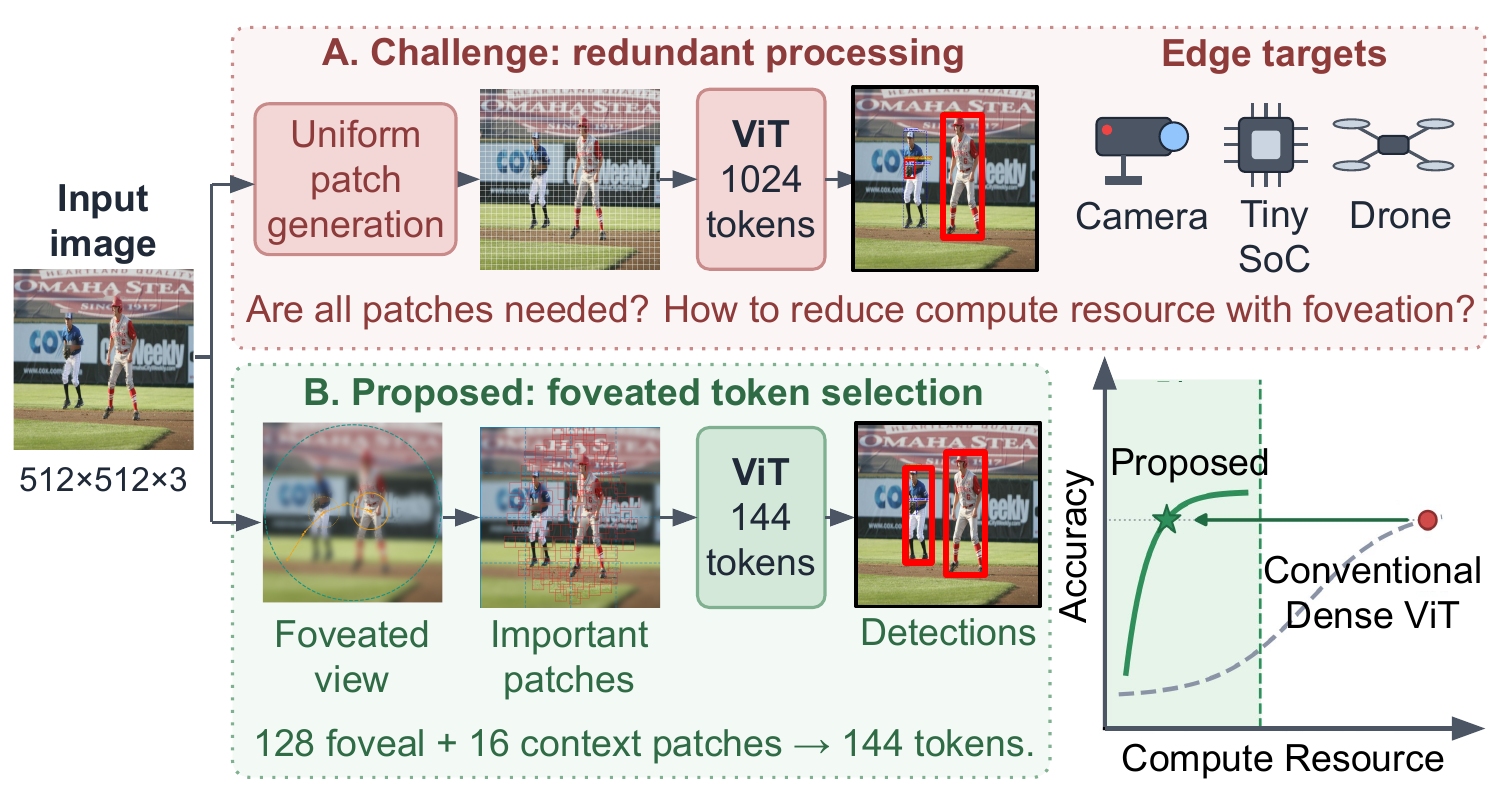}}
\centering
\caption{ A. Challenge of redundant processing and biological counterpart B. Overview of proposed solution}

\label{fig:1}
\end{figure}

Real-time object detection on high-speed mobile autonomous platforms requires operation under stringent latency constraints. However, biologically, not every pixel in an image is equally important. The inefficiency of uniformly embedding patches across the entire image frame is addressed in biological vision systems through a mechanism called foveation (Fig. ~\ref{fig:1}). High-importance salient regions within the image are continuously identified and processed at high resolution, while the periphery is processed at lower resolution. This dynamically allocates computational resources to salient regions. Computationally, this translates to reducing the number of patches in unimportant regions, thus changing the image sampling sparsity for token generation.

Foveation-based image processing has been applied to object detection \cite{GuzmanPando2021DeepFoveaNet}, segmentation \cite{Schmidt2025}, image generation \cite{Chao2026FoveatedDiffusion}, and tracking \cite{Li2017FoveaNet}. Identification of the region of importance or fixation point plays a critical role in the performance of these methods, as it determines the region contributing most to the final decision. Mahajan et al. use entropy calculations over the image frame to estimate the amount of information in different regions for patch generation. However, the salient regions are expected to be present in the four quadrants of the image, an assumption that may not generalize to all object detection approaches \cite{Mahajan2026}. Abid et al. assigns scores to different regions within the frame to select the top contenders for high-resolution processing \cite{Abid2026}. Another approach using dataset statistics from an autonomous driving dataset has been proposed where the long road in the center of the image frame often determines the region of gaze fixation \cite{Thavamani2021}. However, such statistics may be highly correlated to environmental lighting, geography of the location etc. A reinforcement-learning based approach has been proposed in \cite{Ibrayev2024TwoStream} where the gaze of fixation is sequentially moved to different parts of the image before a combined inference is made. Thus, a method with generalizable single-shot prediction of fixation points has not been adequately studied along with hardware evaluation of mapping such approaches on target inference hardware.

To address these shortcomings, we propose a unified CNN–transformer architecture with fixation point identification, token pruning and a detection backend. Our lightweight guidance CNN extracts image features and generates a heatmap to predict important regions. Our token generation method creates patches only around fixation points, enabling up to 65.4\% to 92.2\% token pruning for object detection. We address under-proportional hardware performance by re-engineering parallelizable software kernels for sequential patch generation and developing a RISC-V-IM ISA to accelerate token generation bottleneck. Our software–hardware co-designed approach achieves 4.76$\times$ latency reduction with $-4.75\%$  to $+6.33\%$ mAP degradation.

\section{Software Framework}

\begin{figure}[h!]
\centerline{\includegraphics[width=0.5\textwidth,]{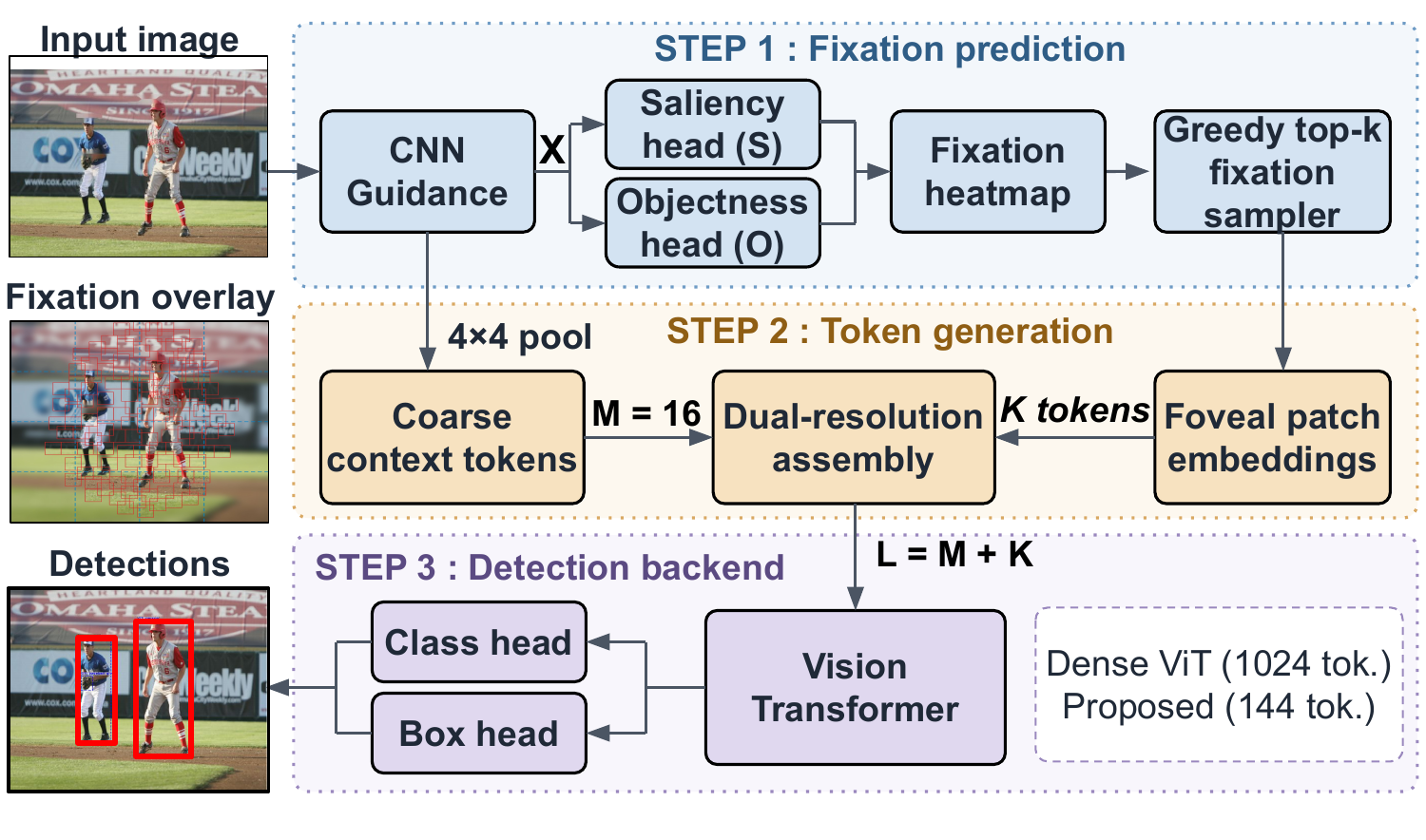}}
\centering
\caption{FoveaDet: (Step 1) A lightweight CNN guidance head generates fixation heatmap to sample $K$ diverse foveal fixations. (Step 2) $K$ high-resolution foveal patches are assembled with coarse global context tokens into a compact sequence. (Step 3) A standard ViT backbone and detection heads predict final class labels and bounding boxes.}

\label{fig:2}
\end{figure}

Fig.~\ref{fig:2} shows the proposed foveated detector with identification of fixation points as the first step. A fixed budget of $K$ high-resolution patches is then sampled at the predicted locations instead of patch generation over the entire image. Only these patches and a few coarse context tokens enter a pretrained ViT. Since selection happens \emph{before} the first transformer block, no compute is spent on discarded background patches and the sequence length is fixed at design time to avoid any irregular patch selection.

\subsection{Saliency and Objectness Guidance}
A lightweight guidance CNN maps the $512{\times}512$ image ($\mathbf{I}$) to a shared feature map $X$ with 48 channels. We carry out the feature map extraction using ConvNeXt-Femto \cite{woo2023convnextv2} with a two-level FPN fusing stride-8 and stride-16 features or MobileNetV3-Small \cite{howard2019searching} at stride 16. Two small convolutional heads are applied to the output of the guidance CNN $X$ - a \emph{saliency} map $S$ and an \emph{objectness} map $O$ in $[0,1]$. They are trained with different targets : $S$ peaks at object centers, while $O$ highlights object boundaries. Therefore, they cover the interior and boundaries together to capture both large-object and small or thin structures. The outputs of $S$ and $O$ are combined in min-max normalization into a fixation heatmap $F$. Specifically, they are normalized into $\hat{S}$ and $\hat{O}$ to compute $F = \alpha \hat{S} + \beta \hat{O}$. The normalization parameters $\alpha $ (=0.6), $\beta$ (=0.4) are empirically selected.

\subsection{Greedy Fixation Sampler}
As the heat map peaks near the objects of interest, we take the $K$ highest values from the heatmap ($F$). However, a naive max-selection may cluster all tokens on a single large object missing other competing objects. Therefore, we use greedy spatial suppression (Algorithm~\ref{alg:sampler}). Once strong heatmap location is selected, every pixel within $d_{\min}=28$\,px window around it is suppressed before the next selection. The selected centers are clamped so that each patch lies inside the image.

\begin{algorithm}[t]
\caption{Foveated Token Generation}
\label{alg:sampler}
\DontPrintSemicolon
\KwIn{Image $\mathbf{I}$; budget $K$; radius $d_{\min}$; weights $\alpha,\beta$}
\KwOut{Token sequence $Z_0$ of length $M{+}K$}
$X,S,O\leftarrow \text{CNN}(\mathbf{I})$\\
$F\leftarrow \alpha\,\text{norm}(S)+\beta\,\text{norm}(O)$\\
\For{$k=1$ \KwTo $K$}{
  $p_k\leftarrow \arg\max F$\\
  $s_k\leftarrow F[p_k]$\\
  $F[\,\|p-p_k\|_2<d_{\min}\,]\leftarrow-\infty$\\
  $p_k\leftarrow \text{clamp}(p_k)$\tcp*{keep patch inside image}
  $e_k\leftarrow \text{PatchEmbed}\big(\text{Crop}_{16\times16}(\mathbf{I};p_k)\big)$\\
  $e_k\leftarrow e_k+\text{PosEnc}(p_k)+\text{ScoreEnc}(s_k)$\;
}
$g_{1..M}\leftarrow \text{Embed}\big(\text{AvgPool}_{4\times4}(X)\big)$\tcp*{coarse context}
\Return{$Z_0=[\,g_1,\dots,g_M,\ e_1,\dots,e_K\,]$}
\end{algorithm}

\subsection{Dual-Resolution Token Sequence}

\textbf{Foveal tokens.} Around each fixation, a $16{\times}16$ patch is cropped at native resolution from the input $\mathbf{I}$ by bilinear grid sampling and embedded with the ViT's own patch-embedding layer. This makes sure that the pretrained weights are reused unchanged. Conventional patching preserves the spatial grid nature of the image by embedding patching is a spatial order. However, foveation fixation centers may not lie at the patch centers of the conventional patching framework. To account for this anomaly, we take a two-step approach. We first carry out bicubic interpolation of the pretrained position embedding at the fixation to estimate the fractional position of a patch between two patches in conventional framework. We then carry out a multi-frequency Fourier Multilayer Perceptron (MLP) of the coordinates with a learnable gain using an MLP embedding of the fixation score $F$. This is similar to the approach of compressed embedding carried out in Neural Radiance Fields (NeRF) \cite{mildenhall2020nerf, tancik2020fourier}. The MLP provides the only differentiable path from the detection loss back into the guidance maps.


\textbf{Context tokens.} The CNN features are average-pooled to a $4{\times}4$ grid and projected to the ViT width, giving $M{=}16$ coarse tokens with learned position embeddings. Therefore, the complete ViT input is a combination of foveal tokens (K) and context tokens (M=16).

\vspace*{-15pt}
\begin{equation}
Z_0=[\,g_1,\dots,g_M,\,e_1,\dots,e_K\,], \quad L=M+K\ll1024 ,
\label{eq:seq}
\end{equation}

\subsection{Fovea-Anchored Detection Head}

The output representations of the $K$ foveal tokens from the transformer backbone are passed to two lightweight prediction heads: a classification MLP (predicting 80 COCO \cite{lin2014microsoft} categories plus background) and a 3-layer box regression MLP. The $M = 16$ coarse context tokens provide global scene awareness during self-attention and are discarded at prediction time. 

Rather than predicting absolute bounding boxes across the entire image canvas from scratch, each foveal token is anchored to its known normalized fixation center, $\mathbf{p}_k = (x_k, y_k) \in [0, 1]^2$. The box MLP only predicts a local center offset $(\Delta x_k, \Delta y_k)$ and scale parameters $(\Delta w_k, \Delta h_k)$ while making sure that the centre stays inside the image.  For a consistent benchmark, the dense baseline applies this identical head across all 1024 patch tokens. 

\begin{figure}[h!]
\centerline{\includegraphics[width=0.5\textwidth]{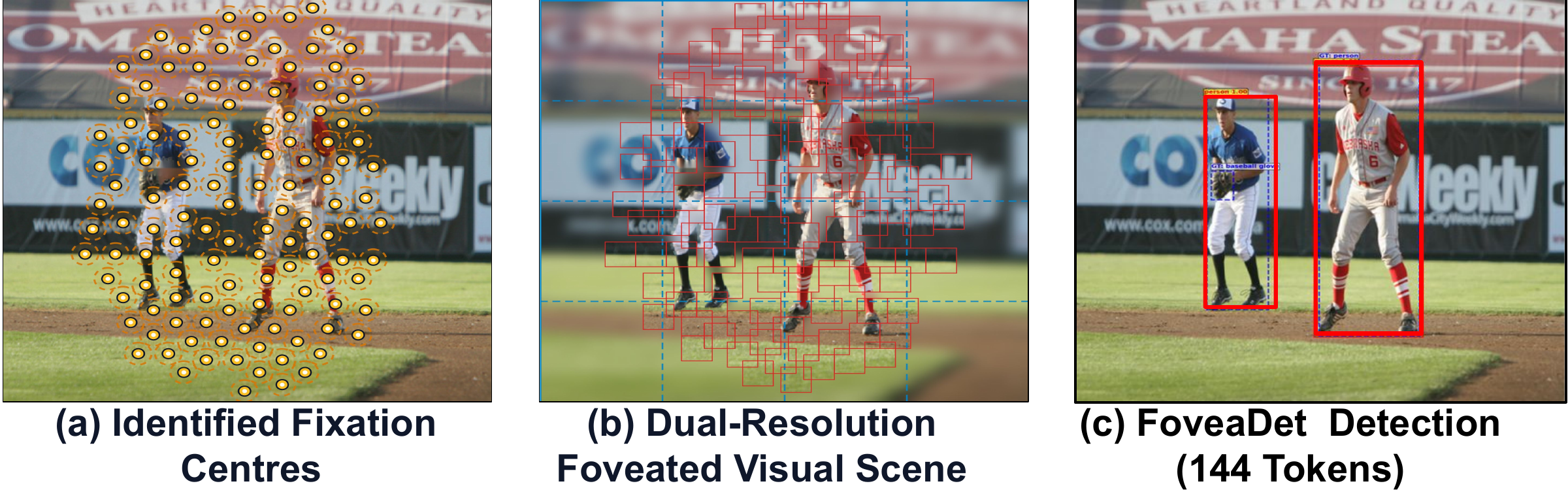}}
\centering
\caption{Example of (a) Identified fixation centres (b) Foveated image (c) FoveaDet detection }
\label{fig:3}
\end{figure}

\begin{table}[t]
\centering
\caption{Performance with ConvNeXt-Femto Guidance}
\label{tab:convnext_results}
\resizebox{\columnwidth}{!}{%
\begin{tabular}{llccc}
\hline
\rowcolor{gray!35}
\textbf{Backbone} & \textbf{$K$} & \textbf{GFLOPs} & \textbf{mAP (\%)} & \textbf{AP$_{50}$ (\%)} \\
\hline
\textbf{DeiT-Distilled} & \cellcolor{gray!15}Dense & \cellcolor{gray!15}10.60 & \cellcolor{gray!15}9.12 & \cellcolor{gray!15}21.07 \\
 & 64  & \cellcolor{savegreen!45}2.28 ($-$78.5\%) & 12.91 & 26.90 \\
 & 128 & \cellcolor{savegreen!40}2.70 ($-$74.5\%) & \textbf{13.95} & \textbf{26.94} \\
 & 256 & \cellcolor{savegreen!25}3.67 ($-$65.4\%) & 11.56 & 23.42 \\
\hline
\textbf{DeiT-Tiny} & \cellcolor{gray!15}Dense & \cellcolor{gray!15}10.60 & \cellcolor{gray!15}8.09 & \cellcolor{gray!15}19.26 \\
 & 64  & \cellcolor{savegreen!45}2.28 ($-$78.5\%) & 11.78 & \textbf{24.23} \\
 & 128 & \cellcolor{savegreen!40}2.70 ($-$74.5\%) & \textbf{12.02} & 23.61 \\
 & 256 & \cellcolor{savegreen!25}3.67 ($-$65.4\%) & 10.50 & 21.27 \\
\hline
\textbf{ViT-Tiny} & \cellcolor{gray!15}Dense & \cellcolor{gray!15}10.60 & \cellcolor{gray!15}6.41 & \cellcolor{gray!15}16.94 \\
 & 64  & \cellcolor{savegreen!45}2.28 ($-$78.5\%) & \textbf{11.06} & \textbf{23.27} \\
 & 128 & \cellcolor{savegreen!40}2.70 ($-$74.5\%) & 10.94 & 23.15 \\
 & 256 & \cellcolor{savegreen!25}3.67 ($-$65.4\%) & 10.85 & 22.50 \\
\hline
\end{tabular}%
}
\end{table}

\begin{table}[t]
\centering
\caption{Performance with MobileNetV3-Small Guidance}
\label{tab:mobilenet_results}
\resizebox{\columnwidth}{!}{%
\begin{tabular}{llccc}
\hline
\rowcolor{gray!35}
\textbf{Backbone} & \textbf{$K$} & \textbf{GFLOPs} & \textbf{mAP (\%)} & \textbf{AP$_{50}$ (\%)} \\
\hline
\textbf{DeiT-Distilled} & \cellcolor{gray!15}Dense & \cellcolor{gray!15}10.60 & \cellcolor{gray!15}9.12 & \cellcolor{gray!15} \textbf{21.07} \\
 & 64  & \cellcolor{savegreen!80}0.83 ($-$92.2\%) & 6.49 & 16.32 \\
 & 128 & \cellcolor{savegreen!70}1.25 ($-$88.2\%) & 8.31 & 19.54 \\
 & 256 & \cellcolor{savegreen!55}2.22 ($-$79.1\%) & \textbf{8.38} & 19.81 \\
\hline
\textbf{DeiT-Tiny} & \cellcolor{gray!15}Dense & \cellcolor{gray!15}10.60 & \cellcolor{gray!15}8.09 & \cellcolor{gray!15}19.26 \\
 & 64  & \cellcolor{savegreen!80}0.83 ($-$92.2\%) & 6.30 & 16.32 \\
 & 128 & \cellcolor{savegreen!70}1.25 ($-$88.2\%) & 8.41 & 19.63 \\
 & 256 & \cellcolor{savegreen!55}2.22 ($-$79.1\%) & \textbf{8.87} & \textbf{20.42} \\
\hline
\textbf{ViT-Tiny} & \cellcolor{gray!15}Dense & \cellcolor{gray!15}10.60 & \cellcolor{gray!15}6.41 & \cellcolor{gray!15}16.94 \\
 & 64  & \cellcolor{savegreen!80}0.83 ($-$92.2\%) & 5.19 & 14.24 \\
 & 128 & \cellcolor{savegreen!70}1.25 ($-$88.2\%) & \textbf{7.01} & \textbf{17.74} \\
 & 256 & \cellcolor{savegreen!55}2.22 ($-$79.1\%) & 6.68 & 16.80 \\
\hline
\end{tabular}%
}
\end{table}

Predictions are assigned to ground-truth objects by Hungarian matching (costs $1/5/2$ for class, L1 and GIoU respectively)\cite{carion2020end}. Inference uses the standard object detection strategies with suppression of detections under a fixed score threshold.

\subsection{Computational Complexity}
In standard ViT, self-attention scales quadratically with patch count:  $\mathcal{O}(N^2 D)$. For a $512{\times}512$ image ($N = 1024$), each layer  computes $1024^2 \approx 1.05 \times 10^6$ pairwise interactions. In FoveaDet with $K = 128$ and $M = 16$, the sequence length is $L = 144$, which requires only $144^2 = 20{,}736$ interactions per layer, with $50.6\times$ reduction in quadratic attention compute. 


\begin{figure}[t]
\centering
\includegraphics[width=\columnwidth]{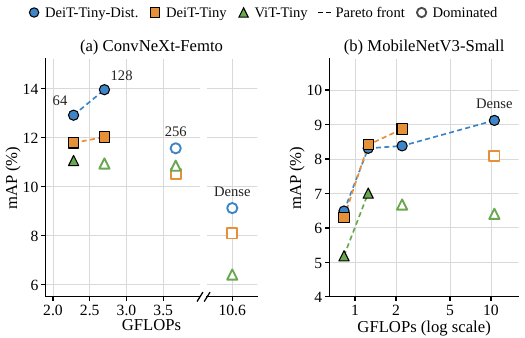}
\caption{Empirical Pareto optimality curve describing the improvement provided by foveated object detection pipeline.}
\label{fig:pareto}
\end{figure}

\section{Experimental Accuracy Evaluation}

We evaluate the proposed foveated object detector on the MS-COCO 2017 dataset at $512{\times}512$ resolution. All models are trained end-to-end for 200 epochs using AdamW and evaluate three lightweight Vision Transformer backbones: \textbf{ViT-Tiny}, \textbf{DeiT-Tiny}, and \textbf{DeiT-Tiny-Distilled}. Two front-end CNN guidance architectures are benchmarked: (i)~\textbf{ConvNeXt-Femto} with multi-scale FPN fusion ($7.93\,\text{M}$ total params), and (ii)~\textbf{MobileNetV3-Small} ($6.04\,\text{M}$ total params), across token budgets $K \in \{64, 128, 256\}$ alongside dense 1024-token baselines ($5.65\,\text{M}$ params). The results are summarized in Table I and Table II.

Most notably, for ConvNeXt-Femto and DeiT-Tiny-Distilled $K{=}128$ variant surpasses the dense baseline by $+4.83\%$ mAP and $+5.87\%$ $\text{AP}_{50}$ with a $74.5\%$ compute reduction. For most other combinations, the mAP suffers from a negligible accuracy degradation with $\sim$90\% suppression in FLOPs. In our prediction, for sub-10M parameter models, processing the entire unpruned $1024$-token grid introduces high spatial redundancy and background noise. By dynamically allocating foveal tokens strictly to salient foreground objects, input-level patch pruning acts as an effective spatial inductive regularizer. A pareto optimal frontier curve describing mitigation of the fundamental GLOPs vs. accuracy trade-off is shown in Fig. 4.

\section{Hardware Framework}
We implemented the proposed algorithmic framework on NVIDIA GTX-1650 GPU to evaluate the token reduction-enabled performance improvement at hardware level. However, we noticed that 65.4\% to 92.2\% FLOP reduction results in much lower performance improvement. We, therefore, carried out meticulous profiling with both software optimization and hardware emulation of specialized vector modules.

\subsection{Software Optimisation Based on Hardware Analysis}

\begin{figure}[h!]
\centerline{\includegraphics[width=0.5\textwidth]{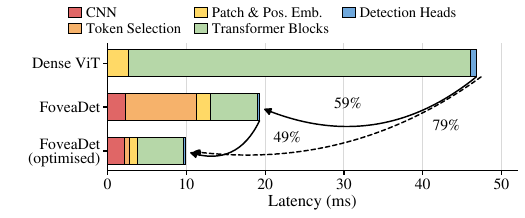}}
\centering
\caption{Stage-wise latency breakdown (ms) of the dense ViT and FoveaDet before and after software optimization on an NVIDIA GTX~1650.}
\label{fig:bar}
\vspace{-3mm}
\end{figure}

Our algorithmic framework demonstrated $\sim50\times$ suppression in operations. However, the hardware speed-up was only $2.4\times$ (46.82\,ms $\rightarrow$ 19.29\,ms) when we deploy the framework naively on a NVIDIA GTX-1650 GPU. Therefore, we carry out mapping and stage-wise profiling to identify bottlenecks. We compared the dense ViT baseline with FoveaDet by dividing the pipeline into five stages: guidance CNN, token selection, patch extraction and embedding, transformer blocks and detection head. The latency of each stage was measured with CUDA events at batch size~1 and $512 \times 512$ resolution, averaged over 200 images after 20 warm-up runs, with both models timed in turn on each image. Our profiling results (Fig. \ref{fig:bar}) shows that token selection accounts for 46.3\% of the total latency of FoveaDet while consuming less than 5\% FLOPs.

We optimize this with two stages in software. The bicubic interpolation of the positional embeddings is rewritten as a single matrix multiplication, cutting it by 89.9\% with numerically identical output. The methods retains over 99\% of the floating-point tokens for both $K=128$ and $K=256$ for 300 COCO images. Together, these optimisations cut end-to-end latency by a further by 79\% (Fig.~\ref{fig:bar}).


\subsection{Algorithm Acceleration on an Emulator}

\begin{figure}[h!]
\centerline{\includegraphics[width=0.5\textwidth]{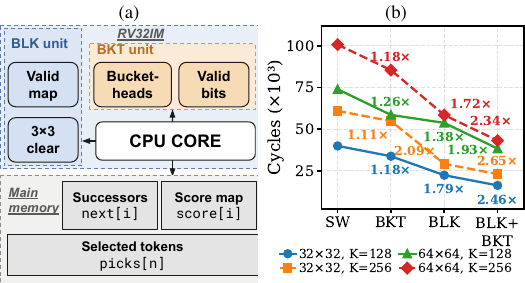}}
\centering
\caption{Block diagram of RISC-V CPU core with token selection acceleration and cycle count for each hardware configuration.}
\label{fig:cycles}
\vspace{-2mm}
\end{figure}

Token selection requires selecting a sorted point within the fixation heatmap while suppressing the points around it. As this process is inherently sequential, it cannot be directly mapped onto a parallel accelerator. Therefore, we propose RISC-V emulation framework for vector operations enabled with custom instructions. The block diagram of our RISCV core supporting RV2IM ISA is based on \cite{Tine2021Vortex} is shown in (Fig.~\ref{fig:cycles}a) with flat memory hierarchy and single cycle execution of instructions. 

Token selection uses a standard bucket sort with two passes. In the first pass, each index $i$ of the quantised score map is inserted into the bucket of its value $v$, where each bucket is a linked list: $\mathit{next}[i] = \mathit{head}[v]$, then $\mathit{head}[v] = i$, with $\mathit{head}[v] = -1$ marking an empty bucket. The BKT unit holds the bucket heads on-core, while the successor array $\mathit{next}$ stays in main memory. In the second pass, the buckets are visited from the highest value down and each cell that is still free is selected until $K$ cells are chosen, and the BLK unit blocks the selected cell's $3\times3$ neighbourhood in a single instruction. Instruction traces show that, in software, blocking a neighbourhood takes a nine-iteration loop and updating the bucket heads in memory takes four of the eleven instructions per cell. Both units are single-cycle and are accessed through \texttt{toksel} instructions written beside the BLK and BKT blocks in (Fig.~\ref{fig:cycles}a). All reductions in  (Fig.~\ref{fig:cycles}b) are relative to the software-only implementation running on the same core.

\begin{table}[t]
\centering
\caption{Comparison with Object Detectors on MS-COCO}
\label{tab:detector_comparison}
\renewcommand{\arraystretch}{1.28} 
\setlength{\tabcolsep}{4.0pt}      
\resizebox{\columnwidth}{!}{%
\begin{tabular}{lccccc}
\hline
\rowcolor{gray!35}
\textbf{Object Detector} & \textbf{Res.} & \textbf{Active Area} & \textbf{Boxes} & \textbf{GFLOPs} & \textbf{mAP (\%)} \\
\hline
\multicolumn{6}{l}{\cellcolor{gray!15}\rule{0pt}{2.5ex}\textit{\textbf{Multi-Scale CNN Detectors} (Feature Pyramids + 1-to-Many)}} \\
\hline
EfficientDet-D0~\cite{tan2020efficientdet} & $512^2$ & 100\% & 49.1k & 2.50 & 33.80 \\
FBNetV5-Det~\cite{wu2021fbnetv5} & $320{\times}640$ & 100\% & 1.0k & 1.80 & 30.40 \\
PP-PicoDet-S~\cite{yu2021picodet} & $416^2$ & 100\% & 3.5k & 1.24 (1.88$^\dagger$) & 30.60 \\
YOLOX-Nano~\cite{ge2021yolox} & $416^2$ & 100\% & 3.5k & 1.08 (1.64$^\dagger$) & 25.30 \\
\hline
\multicolumn{6}{l}{\cellcolor{gray!15}\rule{0pt}{2.5ex}\textit{\textbf{ViT Object Detectors} (Single-Scale Stride-16 + 1-to-1 Hungarian)}} \\
\hline
YOLOS-Tiny~\cite{fang2021yolos} & $512{\times}864$ & 100\% & 100 & 18.80 & 28.70 \\
Dense ViT-Det (Baseline) & $512^2$ & 100\% & 1,024 & 10.60 & 9.12 \\
\hline
\textbf{FoveaDet (MNv3, $K{=}64$)} & $512^2$ & \cellcolor{savegreen!60}\textbf{6.3\%} & \cellcolor{savegreen!60}\textbf{64} & \cellcolor{savegreen!80}\textbf{0.83} & 6.30 \\
\textbf{FoveaDet (MNv3, $K{=}128$)} & $512^2$ & \cellcolor{savegreen!45}\textbf{12.5\%} & \cellcolor{savegreen!45}\textbf{128} & \cellcolor{savegreen!70}\textbf{1.25} & 8.41 \\
\textbf{FoveaDet (MNv3, $K{=}256$)} & $512^2$ & \cellcolor{savegreen!25}\textbf{25.0\%} & \cellcolor{savegreen!25}\textbf{256} & \cellcolor{savegreen!55}2.22 & 8.87 \\
\textbf{FoveaDet (ConvNeXt, $K{=}64$)} & $512^2$ & \cellcolor{savegreen!60}\textbf{6.3\%} & \cellcolor{savegreen!60}\textbf{64} & \cellcolor{savegreen!45}2.28 & 12.91 \\
\textbf{FoveaDet (ConvNeXt, $K{=}128$)} & $512^2$ & \cellcolor{savegreen!45}\textbf{12.5\%} & \cellcolor{savegreen!45}\textbf{128} & \cellcolor{savegreen!40}2.70 & \textbf{13.95} \\
\hline
\multicolumn{6}{l}{\rule{0pt}{2.2ex}\footnotesize $^\dagger$Normalized to identical $512{\times}512$ input resolution (iso-resolution) matching our setup.} \\
\hline
\end{tabular}%
\vspace{-2mm}
}
\end{table}

\section{Discussion}

A comparison of this work with prior literature is shown in Table III. While specialized CNNs achieve higher mAP, FoveaDet deliberately prioritizes edge efficiency. We abandon multi-scale feature pyramids to eliminates the massive memory overhead inherent to high-resolution feature maps~\cite{lin2017feature}. Simultaneously, by predicting only $\le128$ salient bounding boxes instead of $\sim$3500, we bypass the sequential latency bottleneck of Non-Maximum Suppression~\cite{wang2024yolov10}.

Apart from computational optimizations, our three-step method closely follows its biological counterpart where detection centres are often chosen using a bottom-up saliency map \cite{veale2017how} network in V1 primary visual cortex. This results in the allocation of visual resources to familiar objects before the detailed processing in the ventral visual pathway \cite{kravitz2013ventral}.

\section{Conclusion}
We present a software-hardware co-design framework for highly efficient edge object detection. Inspired by biological foveation, our CNN-transformer dynamically identifies fixation points, pruning up to $4\times$ to $16\times$ of background tokens. By combining parallel software optimizations with custom RISC-V IM ISA extensions, this approach achieves a 4.76$\times$ latency reduction with $-4.75\%$  to $+6.33\%$ mAP degradation, advancing next-generation mobile and robotic vision systems.


\section{Acknowledgement}
This work was supported by Seed Grant and Young Faculty Fellowship of IIT Bombay.

\bibliographystyle{ieeetr}
\bibliography{main}

\end{document}